\documentclass{article} 
\usepackage{iclr2027_conference,times}

\usepackage{amsmath,amsfonts,bm}

\def\eqref#1{equation~\ref{#1}}

\def\1{\bm{1}}

\DeclareMathAlphabet{\mathsfit}{\encodingdefault}{\sfdefault}{m}{sl}
\SetMathAlphabet{\mathsfit}{bold}{\encodingdefault}{\sfdefault}{bx}{n}

\usepackage{hyperref}
\usepackage{url}
\usepackage{amsthm}
\usepackage{graphicx}
\usepackage{booktabs}
\usepackage{enumitem}
\newtheorem{proposition}{Proposition}

\title{When Clipping Reverses Correction: Failure Dynamics of Pointwise Forward-KL On-Policy Self-Distillation}

\author{
Di Huang$^{1}$, Hao Li$^{1}$, Yixin Chen$^1$, Fuhai Li$^{1,2*}$\\
$^{1}$Department of Computer Science, $^{2}$Department of Pediatrics\\
Washington University in St. Louis\\
~*Correspondence: \texttt{fuhai.li@wustl.edu}
}

\iclrfinalcopy 
\begin{document}

\maketitle

\begin{abstract}
On-policy self-distillation (OPSD) trains a student on its own generated responses using feedback from the same model conditioned on privileged information. On mathematical reasoning, the original OPSD study finds that stylistic tokens can dominate the training signal over math-related tokens, and that pointwise clipping of the forward KL objective stabilizes training. Pointwise clipping caps each vocabulary-wise forward KL term at a fixed threshold before summing over the vocabulary. Follow-up studies have adopted this clipping, but its effect on training has not been directly examined. In matched training runs differing only in whether clipping is applied, we observe that clipped runs produce substantially more repetitions that persist to the end of the response than their unclipped counterparts. We trace this failure to the clipped objective. We prove that the clipped objective can fail to correct the student toward the teacher and can instead push clipped and unclipped token probabilities away from its teacher. Our training runs agree with this analysis: inside repetitions, the clipped student places less probability than its teacher on leaving the repetition, and more on continuing it, whereas the unclipped runs stay close to their teachers.
\end{abstract}

\section{Introduction}
\label{sec:introduction}

Outcome rewards give a reasoning model one scalar for thousands of tokens. On-policy distillation instead provides a per-token target from a teacher evaluated on the student's own responses \citep{agarwal2024gkd,lu2025opd}. Self-distillation uses the same network as both teacher and student, replacing a stronger teacher model with access to additional privilege information that is withheld from the student \citep{zhao2026opsd,hubotter2026sdpo,shenfeld2026sdft}. Self-distillation provides dense supervision, but how the student learns from it depends on the training objective.

On-Policy Self-Distillation (OPSD) trains the student with a pointwise-clipped forward KL objective \citep{zhao2026opsd}. At each response position, this objective clips each vocabulary-wise forward KL term at a fixed threshold before summing over the vocabulary. The authors introduced this pointwise clipping to keep stylistic tokens from dominating the math-related tokens in the training signal, and many subsequent methods retain it \citep{chen2026dualopsd,li2026unsupervisedopsd,liu2026pwopsd,shrestha2026rethinkingpi}.

Studies using this recipe report response length inflation \citep{yang2026oglssd}, responses that fail to terminate within the generation budget \citep{chen2026dualopsd,ichihara2026opsd}, redundant reasoning chain \citep{gu2026rethinking}, and accuracy that declines after early gains \citep{zhao2026opsd,chen2026dualopsd,ichihara2026opsd,pan2026rlcsd,zhang2026privileged}. Proposed explanations include a frozen teacher that cannot adapt to the student \citep{chen2026dualopsd}, shifts in the teacher's style or reasoning pattern induced by the privileged context
\citep{ichihara2026opsd,pan2026rlcsd,yang2026oglssd,gu2026rethinking}. No study reports that pointwise clipped objective contribute to the text degeneration, leaving this possibility open for further investigation.

Prior work offers a partial mathematical description of this clipped objective: a clipped
term becomes constant and loses its direct gradient \citep{chen2026dualopsd},
and the clipped sum is no longer a divergence
\citep{chen2026dualopsd,feng2026past}. These observations do not establish
which student distributions the clipped objective favors, in particular
whether minimizing it still brings a clipped token's probability closer to the
teacher's.

We trained the pointwise-clipped forward KL OPSD recipe with a frozen reference-conditioned teacher and
encountered a failure mode of text degeneration: generations ended in periodic tails that
never terminated. When we removed only the clipping, nearly all of these loops
disappeared. The teacher and its privileged context are identical in the two runs, so
neither accounts for the difference. We therefore ask whether, and through what
mechanism, this clipped objective contributed to this failure. 

Our contributions therefore follow three questions: how the clipped objective changes the logit gradient,
where its minimum lies, and what it does to training.
\textbf{The logit gradient} (Section~\ref{sec:local}). We show that, relative to exact forward KL, the clipped objective reverses the direction in which
gradient descent moves the logits, on every clipped
token and on unclipped token whose student probability lies within a bounded range
above the teacher's.
\textbf{The minimizer}
(Section~\ref{self-reinforcing}). We show that, at a response position with the
unclipped and clipped token sets held fixed, the clipped objective is minimized only
when the clipped tokens' probability is reduced to zero rather than restored toward
the teacher's, and each unclipped token receives more probability than the teacher
assigns it. A lower total probability on the clipped tokens always permits a lower
objective value.
\textbf{Training runs}
(Sections~\ref{sec:basins}). We train four matched pairs of runs that differ only in whether clipping is applied
and vary whether the student and the teacher think. The clipped runs produce far
more repetitions that continue to the end of the response than
their unclipped runs. Inside repetitions, the clipped student places less
probability than its teacher on leaving, and more on continuing, whereas the unclipped runs
stay close to their teachers. These observations are consistent with the logit-gradient reversals we derived. 

\section{Related Work}
\label{sec:related}

\paragraph{Pointwise-clipped forward KL.}
OPSD introduces the pointwise clipping with threshold  $\tau$ \citep{zhao2026opsd} on each vocabulary-wise forward KL term. Later studies  default threshold $\tau=0.05$ with training methods deviations\citep{tan2026ssopd,hou2026dash,zhang2026privileged,liu2026pwopsd,wang2026trace,chen2026scopeopsd,ichihara2026opsd}, at other values and/or with training methods deviations \citep{chen2026dualopsd,ichihara2026opsd,shrestha2026rethinkingpi,tan2026paint,liang2026searche1}, or without stating one \citep{li2026unsupervisedopsd,zhao2026psopsd}. \citet{chen2026dualopsd} observe that a clipped term no longer supplies a gradient, so the cap stops the student from following large, mostly stylistic terms, and that a sum of capped signed terms is not a divergence. \citet{feng2026past} notes that the clipped objective equals exact forward KL wherever no term exceeds the threshold, so the two share their gradient and Hessian around the student that matches the teacher, but that elsewhere it can be negative.

\paragraph{Failures reported under the clipped objective.}
OPSD reports its best checkpoint, yet its comparison of objectives scores lower at 100 updates than at 50 \citep{zhao2026opsd}. \citet{chen2026dualopsd} finds that its OPSD baseline loses accuracy from 100 to 200 updates while truncation doubles, and also stating frozen teacher that cannot respond to what the student rejects as one of the limitations. \citet{zhang2026privileged} find that their three SmolLM3 students fall below the base model by 100 updates under multiple different configurations. \citet{ichihara2026opsd} finds that, a teacher with a math question and a solution to a physics question leaves many generations repetitive and non-terminating, as compared to one with another math question's solution instead, attribute this to the context without isolating which property is responsible, and leave deterioration under longer training unexplained. \citet{gu2026rethinking} observe that ``OPSD trajectories often recompute the
same intermediate quantities or revise earlier steps without new information,
leading to long and redundant reasoning chains,'' and attribute this to a
teacher conditioned on a single reference solution.

\section{Experimental Setup}
\label{sec:setup}

\paragraph{Training design.}
Our goal is to reproduce the training design of OPSD,
which we reimplement in the verl framework \citep{sheng2025hybridflow}. 
Each run starts from Qwen3-4B \citep{qwen2025qwen3}, with a trainable
full-parameter student and a frozen teacher initialized from the same
checkpoint. The teacher additionally receives a reference solution and provides its
next-token distribution at every position of the student's responses.

We train for 200 updates on the same 30k OpenThoughts \citep{guha2026openthoughts} dataset used to train OPSD. 
We construct
four matched clipped/unclipped pairs. Within each pair, the data, 
sampling setup, and model configuration are identical; only whether 
pointwise clipping is applied differs and we call the unclipped run 
the \emph{twin} of the clipped run. Table~\ref{tab:run-matrix}
summarizes the run-specific configurations. The complete training
hyperparameters and chat template are given in Appendix~\ref{app:training-details}.
We refer to each pair by its letter in Table~\ref{tab:run-matrix}. F, M,
K and L differ only in which side thinks.
The letters index the order in which the configurations entered our run series, each
added to isolate one variable from an earlier run.

\paragraph{Distillation objective.}
OPSD computes its loss over the full vocabulary. Our teacher runs as a
separate inference service under SGLang \citep{zheng2024sglang}, and
sending a full-vocabulary distribution for every response position is
impractical, so our interface returns the teacher's top-128 token log
probabilities at each response position.

At response position $t$, let support $\mathcal S_t$ denote set of
the teacher's top-128 tokens, and
let $a_{t,i}$ and $b_{t,i}$ be the teacher's and the student's logits for
token $i$. Both distributions are formed by a softmax over $\mathcal S_t$
at temperature $T=1.1$,
\begin{equation}
p_{t,i}=\frac{\exp(a_{t,i}/T)}{\sum_{j\in\mathcal S_t}\exp(a_{t,j}/T)},
\qquad
q_{t,i}=\frac{\exp(b_{t,i}/T)}{\sum_{j\in\mathcal S_t}\exp(b_{t,j}/T)},
\qquad i\in\mathcal S_t,
\label{eq:support-distributions}
\end{equation}
so that the teacher distribution $p_t$ and the student distribution $q_t$
each sum to one on $\mathcal S_t$. We then define the vocabulary-wise forward KL term of token $i$:
\begin{equation}
d_{t,i}=p_{t,i}\log\frac{p_{t,i}}{q_{t,i}}.
\label{eq:forward-kl-contribution}
\end{equation}
Here $d_{t,i}$ is positive where the student assigns token $i$ less
probability than the teacher and negative where it assigns more. Exact
forward KL sums $d_{t,i}$ directly, whereas the pointwise-clipped forward KL
of OPSD first replaces any $d_{t,i}$ above $\tau$ by $\tau$, leaving smaller
and negative $d_{t,i}$ unchanged:
\begin{equation}
D_{\mathrm{FKL},t}=D_{\mathrm{KL}}(p_t\Vert q_t)
= \sum_{j\in\mathcal S_t} d_{t,j},
\qquad
D_{\mathrm{clip},t}
= \sum_{j\in\mathcal S_t}\min(d_{t,j},\tau),
\qquad
\tau=0.05.
\label{eq:losses}
\end{equation}
Despite the symbol, which follows OPSD's notation, $D_{\mathrm{clip},t}$ is
not a divergence, since it can be negative. The training loss is the mean of
$D_{\mathrm{clip},t}$ in the clipped runs, or of $D_{\mathrm{FKL},t}$ in the
unclipped runs, over all valid response positions in the update. We call the
tokens in the teacher support $\mathcal S_t$ the \emph{candidate tokens} at 
position $t$, to distinguish them from the token the student emits there, 
and a candidate token with $d_{t,i}>\tau$ an \emph{over-threshold} token. 
In a clipped run, the over-threshold tokens are the clipped tokens.

\begin{table}[t]
\centering
\small
\caption{Paired training families. All use Qwen3-4B, teacher top-128 support, $T=1.1$, 30 problems per update with one rollout per problem, learning rate of $1\times10^{-6}$, forward-KL distillation, and one trajectory per run. Clipped runs use $\tau=0.05$ and their unclipped runs do not.}
\label{tab:run-matrix}
\begin{tabular}{@{}lccc@{}}
\toprule
Family & Reference & Student/teacher thinking & Response cap \\
\midrule
F & yes & off/on & 18,432 \\
M & yes & on/off & 18,432 \\
K & yes & off/off & 8,192 \\
L & yes & on/on & 18,432 \\
\bottomrule
\end{tabular}
\end{table}

\paragraph{Trajectory analysis and evaluation.}
At every response position of every update, we record the teacher's and
the student's log probabilities on the candidate tokens.
We can therefore compute both the clipped objective
and exact forward KL at the same response positions in both runs of each pair.
Full capture details are provided in Appendix~\ref{app:capture}.

We evaluate checkpoints saved every 25 training updates on AIME 2025 \citep{maa2026aime} and
AIME 2024 \citep{aime24}. The main text reports AIME 2025 only, with two metrics:
per-sample accuracy (avg@12) and the terminal loop rate (generation ended in periodic tails that
never terminated). 
The complete AIME 2024 and AIME 2025 results, and evaluation configuration are given in Appendix~\ref{app:evaluation}.

\section{Failure Dynamics of Pointwise Clipping}
\label{sec:mechanism}

\subsection{Pointwise clipping reverses the logit gradient}
\label{sec:local}

For an over-threshold candidate token, the exact forward KL and the pointwise-clipped
forward KL produce gradients on that token's logit with opposite signs.
Exact KL gives a negative gradient, whereas the clipped objective gives a positive gradient.
For a candidate token that is
not over-threshold, the two gradients again have opposite signs exactly when the
student's probability exceeds the teacher's but stays below the teacher's probability
divided by the teacher's total probability on the tokens that are not over-threshold:
exact KL then gives a positive gradient, whereas the clipped objective gives a negative one.

Fix one response position and suppress the index $t$. The teacher is
frozen, so $p$ and its support $\mathcal S$ do not depend on the student's
logits. We use $d_i=p_i\log(p_i/q_i)$ from
Equation~\ref{eq:forward-kl-contribution}. Let $z_i=b_i/T$ be the
temperature-scaled student logit, so that
$q_i=\exp(z_i)/\sum_{j\in\mathcal S}\exp(z_j)$.
We call the derivative with respect to $z_i$ as
its \emph{logit gradient} for candidate token $i$. Define
\[
A=\{j\in\mathcal S:d_j\le\tau\},
\qquad
C=\mathcal S\setminus A,
\qquad
P_A=\sum_{j\in A}p_j.
\]
Here, $A$ and $C$ are the active and clipped candidate token sets, and $P_A$
is the teacher's total probability mass on the active set.
Equation~\ref{eq:losses} becomes
\begin{equation}
D_{\mathrm{clip}}
=\sum_{j\in A}p_j\log\frac{p_j}{q_j}+|C|\tau.
\label{eq:clip-region}
\end{equation}

\begin{proposition}[Logit gradient reversal on clipped candidate tokens]
\label{prop:piecewise-gradient}
\normalfont
For any threshold $\tau>0$ and any student distribution with
$d_j\neq\tau$ for every $j\in\mathcal S$, the gradient of
$D_{\mathrm{clip}}$ with respect to $z_i$ is
\begin{equation}
\frac{\partial D_{\mathrm{clip}}}{\partial z_i}
=P_Aq_i-p_i\mathbf{1}\{i\in A\}.
\label{eq:piecewise-gradient}
\end{equation}
Every clipped candidate token $i\in C$ satisfies
\begin{equation}
\frac{\partial D_{\mathrm{FKL}}}{\partial z_i}
=q_i-p_i<0,
\qquad
\frac{\partial D_{\mathrm{clip}}}{\partial z_i}
=P_Aq_i>0.
\label{eq:logit-gradient-reversal}
\end{equation}
\end{proposition}

Appendix \ref{app:proofs} gives the proof. Both inequalities in Equation~\ref{eq:logit-gradient-reversal}
are strict, and clipping is what makes their gradient signs differ. For
a clipped candidate token \(i\in C\), \(d_i>\tau>0\) implies \(p_i>q_i\), so the
exact KL logit gradient \(q_i-p_i\) is negative. Pointwise clipping
instead caps the term \(d_j\) of every \(j\in C\) at the constant
\(\tau\), which removes the gradient term \(-p_i\) in Equation~\ref{eq:piecewise-gradient}.
The remaining gradient \(P_Aq_i\) is positive because \(q_i>0\)
under softmax and \(P_A>0\). Since \(p\) and \(q\) are both normalized over the same
support, there exists some \(j\) with \(q_j\ge p_j\). Hence \(d_j\le0<\tau\), so
\(j\in A\), and therefore \(P_A\ge p_j>0\), since
Equation~\ref{eq:support-distributions} gives \(p_j>0\).

Clipping also changes the logit gradient on the active candidate tokens when the
student probability lies within a bounded range above the teacher's.
When the clipped set is nonempty, $P_A<1$, so $p_i/P_A>p_i$, for $i\in A$.
By equation \ref{eq:piecewise-gradient}, if the student's probability in the active set satisfies
\[
p_i<q_i<p_i/P_A,
\]
the exact KL gradient $q_i-p_i$ is positive while the clipped gradient \(P_Aq_i-p_i < 0\) is negative.

\subsection{Clipped objective moves probability from clipped to active tokens}
\label{self-reinforcing}
Proposition~\ref{prop:piecewise-gradient} shows that, for a clipped candidate
token, which already has less student than the teacher probability, the logit
gradient of the clipped objective points toward lowering its logit, whereas
that of exact KL points toward raising it. This reversal does not by itself
determine whether the token's probability is still restored toward the teacher's,
as under exact KL, because softmax probabilities depend on all logits.
We therefore consider the
clipped candidate tokens jointly and ask what happens at the minimizer of the
clipped objective at a fixed response position: does an optimized student favor restoring the probability toward the teacher's, or reduced even further?

At a fixed token position, for fixed nonempty active and clipped sets $A$ and $C$, we consider the
constrained problem:
\[
\begin{aligned}
\underset{q}{\operatorname{minimize}}\quad
& D_{\mathrm{clip}}(q)=\sum_{j\in\mathcal S}\min(d_j,\tau)\\
\text{subject to}\quad
& q_i\ge0\ \text{ for every } i\in\mathcal S,
  \qquad \sum_{j\in\mathcal S}q_j=1,\\
& d_i\le\tau\ \text{ for every } i\in A,
  \qquad d_i>\tau\ \text{ for every } i\in C .
\end{aligned}
\]
The feasible set of student distributions, denoted the region $R$, is the
\emph{fixed clipping region} of $(A,C)$. Unlike in Section~\ref{sec:local},
where $q$ is a softmax of finite logits and every $q_i>0$,
we allow $q_i=0$ so that a minimizer exists. For such a candidate token $i$,
$d_i=+\infty$, the capped term is still $\tau$, so it
stays clipped. Everywhere in $R$, $D_{\mathrm{clip}}$ is given
by Equation~\ref{eq:clip-region} with these fixed sets.

\begin{proposition}[Fixed-region minimizer of the clipped objective]
\label{prop:support-pruned}
\normalfont
The clipped objective has a unique minimizer over the fixed clipping
region $R$,
\begin{equation}
q_i^\star=
\begin{cases}
p_i/P_A, & i\in A,\\
0, & i\in C,
\end{cases}
\qquad
D_{\mathrm{clip}}(q^\star)=P_A\log P_A+|C|\tau .
\label{eq:fixed-region-minimizer}
\end{equation}
Let $Q_C=\sum_{j\in C}q_j$ be the student's total probability mass
on the clipped set. For every value of $Q_C$ attained in $R$, if its value is further added to the constraints without inconsistencies with the other constraints, the minimum of
$D_{\mathrm{clip}}$ over the distributions of $q$'s in $R(Q_C)$ with that value is
strictly increasing in $Q_C$:
\begin{equation}
D_{\mathrm{clip}}^\star(Q_C)
=P_A\log P_A-P_A\log(1-Q_C)+|C|\tau,
\qquad
\frac{dD_{\mathrm{clip}}^\star}{dQ_C}=\frac{P_A}{1-Q_C}>0 .
\label{eq:support-pruned}
\end{equation}
\end{proposition}

Appendix~\ref{app:proof-prop2} gives the proof. At the minimizer $q^\star$
of the clipped objective, the clipped candidate tokens' total probability is not
restored toward the teacher's total $1-P_A$ on these tokens but reduced
further, to zero. All probability then lies on $A$, and the student's probability
on each active candidate token equals the teacher's probability $p_i$ multiplied
by the same factor $1/P_A>1$. Clipping
therefore does more than limit the influence of dominant terms: inside a
fixed clipping region, its minimizer assigns zero student probability to
clipped candidate tokens, even though they already have less student probability than
teacher probability, and gives each active candidate token more student
probability than the teacher probability.

A softmax of finite logits, however, never attains $q^\star$, because it
gives every clipped token positive probability. Hence $Q_C>0$, and
$D_{\mathrm{clip}}(q^\star)$ is an unattained infimum. For such a student,
Equation~\ref{eq:support-pruned} characterizes the minimum attainable
objective value for each $Q_C$. The lower the student's total probability
on the clipped candidate tokens, the closer the
objective can come to this infimum, and restoring $Q_C$ toward the teacher's
$1-P_A$ only moves this minimum attainable value farther from it.

\subsection{Clipped updates turn started repetition into persistent loops}
\label{sec:basins}
\begin{figure}[t]
\centering
\includegraphics[width=\linewidth]{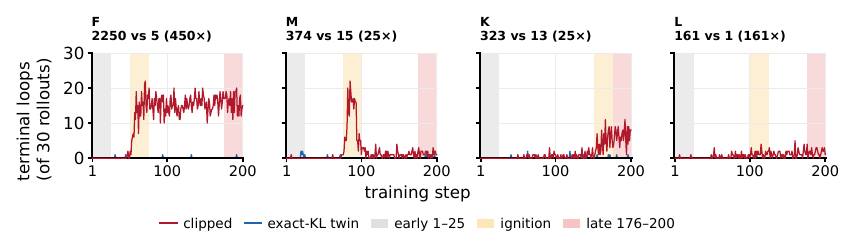}
\caption{Loop outcome over training, one trajectory per run. Each point is the
number of the 30 rollouts sampled at that update that end in a terminal loop.
Clipped runs in red, unclipped twins in blue. Shaded windows are three phases:
early (updates 1--25, gray), ignition (F 51--75, M 76--100,
K 151--175, L 101--125, orange), and late (updates 176--200, pink).}
\label{fig:loop-outcome}
\end{figure}

In matched runs that differ only in whether the objective is clipped, clipped training
produces substantially more terminal loops than exact KL training
(Figure~\ref{fig:loop-outcome}; Table~\ref{tab:persistence-progression}).
We defined the repetition-related terminologies in Table~\ref{tab:loop-terms} and 
use them throughout the section. 
The preceding two sections give a candidate cause: the
clipped objective reverses the logit gradient of exact KL on every clipped token
and on active tokens within a bounded range above the teacher's probability.
Within a fixed clipping region, its minimizer sets the clipped tokens' probability
to zero and gives each active token more probability than the teacher, instead of
correcting the student toward the teacher. If, in training, the teacher's exit 
tokens are clipped and the copy token lies in this range, the clipped objective 
would push the student to leave a repetition less
often and to continue it more often than the teacher, so repetitions and terminal loops 
would be more than in the exact KL twin, although the size of this
effect may differ across families.
Both results describe one response position with a given clipped set, whereas
in training the clipped set changes across sampled responses and positions, so
whether the student's probabilities move this way must be measured. We compare
student and teacher exit mass at the first copy, which every repetition passes
through, and the copy token's logit gradient under both objectives before
maturity. These comparisons can support the logit gradient reversal as the path
from clipping to loops but cannot establish that path.

\begin{table}[!ht]
\centering
\footnotesize

\caption{Definitions used to characterize token repetition and
analyze its continuation. The upper block defines periodic segments,
repetitions, mature repetitions, and terminal loops. The lower block
defines the first copy, copy token, and exit mass.}
\label{tab:loop-terms}
\begin{tabular}{@{}p{1.1in}p{4.2in}@{}}
\toprule
Term & Definition \\
\midrule
Periodic segment & A maximal contiguous sequence $x_s,\ldots,x_e$,
  where $s$ and $e$ are its start and end positions, with period $\ell$
  and length $L=e-s+1>\ell$. $L$ satisfies $x_t=x_{t-\ell}$ for
  $s+\ell\le t\le e$.  \\
Repetition & A periodic segment with at least two complete period
  $\lfloor L/l\rfloor \ge 2$. \\
Mature repetition & A repetition $\lfloor L/l\rfloor \geq 3$ with $L \ge 30$ tokens. \\
Terminal loop & Mature repetition that reaches the last response token. \\
\midrule
Copy token & At a position $t\ge s+\ell$ while the periodic
  continuation remains, the token $c_t=x_{t-\ell}$
  that continues the period. Its teacher and student probabilities are
  $p_{t,\mathrm{copy}}$ and $q_{t,\mathrm{copy}}$, respectively. \\
First copy & Copy token belong to the second completed occurrence of the period,
  at positions $s+\ell,\ldots,s+2\ell-1$. \\
Exit mass & At position $t$, the probability assigned to tokens
  other than the copy token: $1-p_{t,\mathrm{copy}}$ for the teacher
  and $1-q_{t,\mathrm{copy}}$ for the student. \\
\bottomrule
\end{tabular}
\end{table}

\paragraph{Outcome.}
We define three 25-update phases: an early phase at the start of training, an
ignition window at each clipped run's loop onset, and a late phase at
the end of training. In the early phase, terminal loops are nearly absent from every
run. The exact KL runs remain there for the whole run, never exceeding 2
of 30 rollouts at any update. The clipped runs do not. Each run has a
family-specific onset after which terminal loops recur,
and over the full run it produces 25 to 450 times as many of them as its
twin. The result is shown in Figure~\ref{fig:loop-outcome}.
The thinking-mode configuration sets both the scale and
the ignition time.

\paragraph{Trend.}
Table~\ref{tab:persistence-progression} follows repetitions ($N\geq2$)
through three phases. In the early phase the two runs of every pair are
indistinguishable except run L. They generate repetition and mature repetition at the same rate.
From ignition on, two amplifications hold. First, a
repetition grows into a mature repetition ($N\geq3$ and $L\geq30$) more often
under clipping than in the twin. The count $N(\text{mat.}\mid N\geq 2)$ is
8.6 to over 400 times the twin's, and the rate
$P(\text{mat.}\mid N\geq 2)$ 4.5 to over 150 times. Second, far
more repetitions end as terminal loops: 26 to 381 per phase in the clipped
runs, against at most two in any twin.

\paragraph{Family-specific loop patterns.}
\label{par:family-specific-loop-patterns}
With student thinking off and teacher thinking on (F), the student writes
answer-mode derivations that the teacher scores where its own thinking block
would begin. The student chains the repeated ``$\Rightarrow$'' where the teacher prefers \texttt{\textbackslash text}. This two-token cycle supplies most mature
repetitions at ignition, and by the late phase all of them are terminal loops.
With both modes off (K), student and teacher share the answer register, and most of its
terminal loops are answer-closing formatting such as nested ``\textbackslash boxed\{''.
The teacher prefers a content
command such as \texttt{\textbackslash text} or \texttt{\textbackslash frac}. 
With both modes on (L), most mature repetitions lie inside
the thinking block and repeat an equation or a reasoning frame such as ``Let me
compute:'', where the teacher's preferred alternative varies, most often a space,
an opening parenthesis, or a different word such as ``take''. Most of its mature
repetitions exit rather than become terminal. With student thinking on and teacher
thinking off (M), the teacher's prompt contains a closed, empty thinking block, so
the teacher scores the student's thinking as answer text, and both runs stop
emitting the closing tag. At ignition, M's terminal loops repeat an answer-closing
mark such as a check mark inside the unclosed thinking block, where the teacher
prefers the end-of-response token \texttt{<|im\_end|>}.

\begin{figure}[t]
\centering
\includegraphics[width=\linewidth]{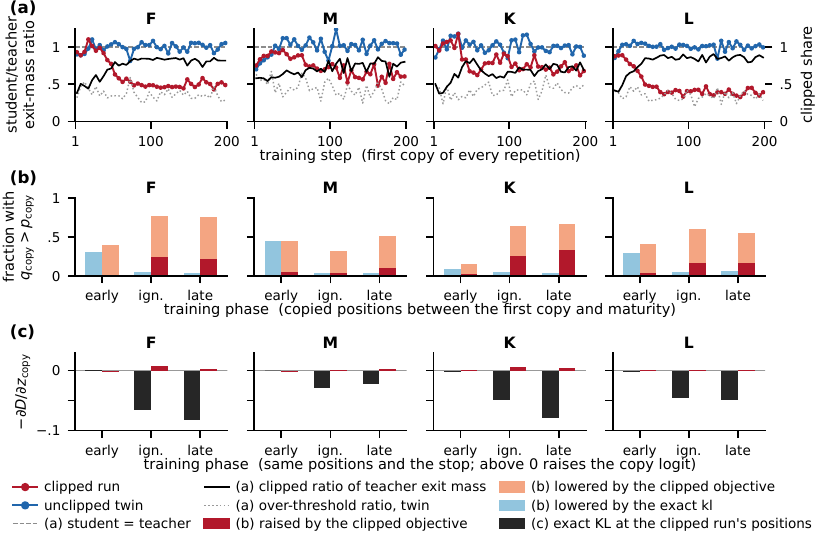}
\caption{The copy token inside repetitions (period 2--50), with the student's
probabilities taken before the sampler's truncation. (a) Student/teacher
exit-mass ratio, one point per block of five updates
(1--5, \ldots, 196--200). Lines without markers give the ratio of the
teacher's exit mass on clipped tokens, or on over-threshold tokens in the twins
(right axis).
(b) Copied positions between the first copy and maturity: fraction
with $q_{\mathrm{copy}}>p_{\mathrm{copy}}$.
Dark bar: $q_{\mathrm{copy}}<p_{\mathrm{copy}}/P_A$, where the logit gradient of
exact KL points toward lowering the copy logit and that of the clipped
objective toward raising it. Light bar: both point toward lowering it.
(c) The same copied positions, plus observed stopping positions
before maturity: mean $-\partial D/\partial z_{\mathrm{copy}}$.
Black and red: exact KL (hypothetical) and clipped KL, respectively,
evaluated on the same positions and probabilities from
the clipped run. Positive values point toward raising
the copy logit. Phases as in Table~\ref{tab:persistence-progression}.}
\label{fig:copy-gradient}
\end{figure}

\paragraph{Exit mass at the first copy.}
We dive deep into the training signal in each runs and how clipped objective applied to it.
We examine whether clipping suppresses exit mass at the first copy
of a repetition, terms defined in Table~\ref{tab:loop-terms}.
Let $\mathcal F_u$ denotes
the window containing first-copy positions of all detected repetitions with periods
$2\le\ell\le50$ across responses generated at update $u$.
For each nonoverlapping five-update block $B$, we compute the
student/teacher exit-mass ratio by summing each distribution's
exit mass over these positions:
\[
R_B=
\frac{\sum_{u\in B}\sum_{t\in\mathcal F_u}
(1-q_{u,t,\mathrm{copy}})}
{\sum_{u\in B}\sum_{t\in\mathcal F_u}
(1-p_{u,t,\mathrm{copy}})}.
\]
We then measure the clipped ratio of the
teacher's exit mass that lies on clipped tokens. For the unclipped twins, we report the
corresponding fraction on over-threshold tokens
(Figure~\ref{fig:copy-gradient}a).
Pooled over the early phase, the student/teacher exit-mass ratio is .79 to 1.04.
Under exact KL, the ratio is close to 1 across the whole run,
even though 33 to 45 percent of the teacher's exit mass
still lies on over-threshold tokens. Under clipping, however, the exit mass ratio falls to
.41 to .70 in the late phase, while 71 to 84 percent of the teacher's exit mass is
clipped. Thus, at the first copy, the clipped student places less probability than its teacher on leaving
the repetition, and most of the teacher's exit mass lies on the tokens whose logit gradient
Proposition~\ref{prop:piecewise-gradient} reverses.

\begin{table*}[t]
\centering
\small
\caption{Progression of repetition by training phase (period 2--50). Rows: $n(N \geq 2)$, the number of repetitions; $N(\text{mat.} \mid N \geq 2)$, the mature repetitions among repetitions; $P(\text{mat.}\mid N\geq 2)$
the proportion of repetitions that mature; $N(\text{term.} \mid \text{mat.})$, the terminal loops among mature repetition. Each cell reads early $\to$ ignition $\to$ late, the three phases of Figure~\ref{fig:loop-outcome}, 750 rollouts each.}
\label{tab:persistence-progression}
\scriptsize

\begin{tabular*}{\linewidth}{@{\extracolsep{\fill}}llrrrr@{}}
\toprule
Quantity & Arm & F & M & K & L \\
\midrule
$n(N\geq 2)$ & clipped & $1195\!{\scriptscriptstyle\to}\!4954\!{\scriptscriptstyle\to}\!2814$ & $3852\!{\scriptscriptstyle\to}\!2666\!{\scriptscriptstyle\to}\!1304$ & $1163\!{\scriptscriptstyle\to}\!3757\!{\scriptscriptstyle\to}\!4024$ & $5952\!{\scriptscriptstyle\to}\!5704\!{\scriptscriptstyle\to}\!6374$ \\
 & twin & $1237\!{\scriptscriptstyle\to}\!2395\!{\scriptscriptstyle\to}\!3154$ & $3288\!{\scriptscriptstyle\to}\!975\!{\scriptscriptstyle\to}\!1496$ & $1130\!{\scriptscriptstyle\to}\!1042\!{\scriptscriptstyle\to}\!1043$ & $3991\!{\scriptscriptstyle\to}\!2642\!{\scriptscriptstyle\to}\!3121$ \\
\midrule
$N(\text{mat.}\mid N\geq 2)$ & clipped & $1\!{\scriptscriptstyle\to}\!790\!{\scriptscriptstyle\to}\!749$ & $6\!{\scriptscriptstyle\to}\!409\!{\scriptscriptstyle\to}\!69$ & $4\!{\scriptscriptstyle\to}\!170\!{\scriptscriptstyle\to}\!176$ & $16\!{\scriptscriptstyle\to}\!195\!{\scriptscriptstyle\to}\!328$ \\
 & twin & $5\!{\scriptscriptstyle\to}\!3\!{\scriptscriptstyle\to}\!7$ & $16\!{\scriptscriptstyle\to}\!1\!{\scriptscriptstyle\to}\!8$ & $6\!{\scriptscriptstyle\to}\!4\!{\scriptscriptstyle\to}\!8$ & $7\!{\scriptscriptstyle\to}\!20\!{\scriptscriptstyle\to}\!9$ \\
\midrule
$P(\text{mat.}\mid N\geq 2)$ & clipped & $.001\!{\scriptscriptstyle\to}\!.159\!{\scriptscriptstyle\to}\!.266$ & $.002\!{\scriptscriptstyle\to}\!.153\!{\scriptscriptstyle\to}\!.053$ & $.003\!{\scriptscriptstyle\to}\!.045\!{\scriptscriptstyle\to}\!.044$ & $.003\!{\scriptscriptstyle\to}\!.034\!{\scriptscriptstyle\to}\!.051$ \\
 & twin & $.004\!{\scriptscriptstyle\to}\!.001\!{\scriptscriptstyle\to}\!.002$ & $.005\!{\scriptscriptstyle\to}\!.001\!{\scriptscriptstyle\to}\!.005$ & $.005\!{\scriptscriptstyle\to}\!.004\!{\scriptscriptstyle\to}\!.008$ & $.002\!{\scriptscriptstyle\to}\!.008\!{\scriptscriptstyle\to}\!.003$ \\
\midrule
$N(\text{term.}\mid\text{mat.})$ & clipped & $0\!{\scriptscriptstyle\to}\!307\!{\scriptscriptstyle\to}\!381$ & $1\!{\scriptscriptstyle\to}\!206\!{\scriptscriptstyle\to}\!26$ & $0\!{\scriptscriptstyle\to}\!109\!{\scriptscriptstyle\to}\!155$ & $1\!{\scriptscriptstyle\to}\!30\!{\scriptscriptstyle\to}\!35$ \\
 & twin & $0\!{\scriptscriptstyle\to}\!0\!{\scriptscriptstyle\to}\!1$ & $6\!{\scriptscriptstyle\to}\!0\!{\scriptscriptstyle\to}\!1$ & $0\!{\scriptscriptstyle\to}\!1\!{\scriptscriptstyle\to}\!2$ & $0\!{\scriptscriptstyle\to}\!1\!{\scriptscriptstyle\to}\!0$ \\
\bottomrule
\end{tabular*}
\end{table*}

\paragraph{Copy token probability before maturity.}
We next examine the positions a repetition passes through before it becomes mature: how often the student places more probability than the teacher on the copy token, and whether each objective corrects this excess.

For each 25-update phase $\phi$, let $\mathcal G_\phi$ denote the window of
copied positions after the first copy and before maturity or an
observed stop.
We record the fraction at which the student assigns
greater probability to the copy token than the teacher:
\[
R_\phi=
\frac{\sum_{t\in\mathcal G_\phi}
\mathbf{1}\{q_{t,\mathrm{copy}}>p_{t,\mathrm{copy}}\}}
{|\mathcal G_\phi|},
\]
where $\mathbf{1}\{\cdot\}$ is the indicator function and counts
are pooled across repetitions and updates within the phase.
The result is in Figure~\ref{fig:copy-gradient}b.
In the early phase, both runs of F, M, and L have
$q_{\mathrm{copy}}>p_{\mathrm{copy}}$ at 29 to 45 percent of these positions.
After ignition, this fraction falls to 4 to 6 percent in the exact KL runs but
32 to 77 percent in the clipped runs. In the clipped runs,
$q_{\mathrm{copy}}$ lies specifically between $p_{\mathrm{copy}}$ and
$p_{\mathrm{copy}}/P_A$ at 4 to 34 percent of positions.
The dark part of each bar in Figure~\ref{fig:copy-gradient}b shows the fraction of positions at which the copy token's logit
gradient is reversed: the clipped objective points toward raising the copy token
logit where exact KL points toward lowering it.

We then measure the gradient of each objective
$D\in\{D_{\mathrm{FKL}},D_{\mathrm{clip}}\}$ on the copy token.
For each phase $\phi$, we record the mean of $-\partial D/\partial z_{\mathrm{copy}}$
over the window $\mathcal G_\phi$ plus the observed stop positions, pooled across repetitions and updates.
We report the negative of the gradient so that positive
values point toward raising the copy token logit.
We calculate
$\partial D_{\mathrm{FKL}}/\partial z_{\mathrm{copy}}=q_{\mathrm{copy}}-p_{\mathrm{copy}}$
and $\partial D_{\mathrm{clip}}/\partial z_{\mathrm{copy}}$, given by
Equation~\ref{eq:piecewise-gradient}. We evaluate both objectives on the positions and probabilities of the clipped run only, so that any difference between the two means comes from the objective alone.
Figure~\ref{fig:copy-gradient}c shows the result: the net direction of each
objective's logit gradient on the copy token over all positions of the window.
In the early phase, both values are nearly zero in every family.
In the ignition and late phases, the mean of $-\partial D_{\mathrm{FKL}}/\partial z_{\mathrm{copy}}$
is negative, pointing toward lowering the copy token logit, whereas that of
$D_{\mathrm{clip}}$ is slightly positive in every family, pointing toward raising it.
At the positions that
decide whether a repetition becomes mature, exact KL would thus correct the student's
excess probability on the copy token, and the clipped objective removes this correction.

\section{Held-Out Evaluation}
\label{sec:evaluation}

\begin{figure}[!t]
\centering
\includegraphics[width=\linewidth]{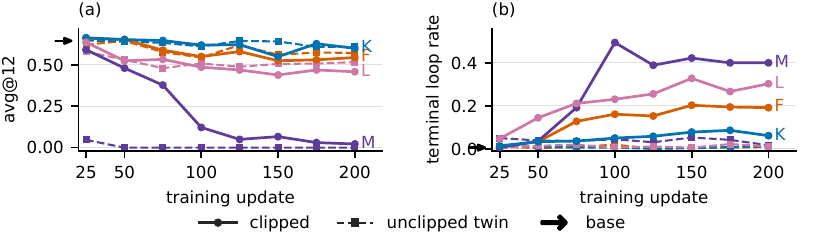}
\caption{AIME~2025 across training. (a) Avg@12, the mean correct ratio over 12 samples for each of 30 problems at the checkpoint saved after each 25 updates. (b) The ratio of the 360 responses per checkpoint that end in a terminal loop. Solid curves with circles are the clipped runs, dashed curves with squares their unclipped twins. Each trained curve is one realized trajectory. The black arrow on each vertical axis marks the base model, the fixed starting checkpoint averaged over eight evaluation-sampling seeds.}
\label{fig:aime25-row}
\end{figure}

We evaluate every checkpoint, saved at 25-update intervals, on the
AIME~2025 problems. Figure~\ref{fig:aime25-row}
reports accuracy and terminal loop rate at saved checkpoints. It tests
whether clipping increases looping and whether the additional loops
accompany accuracy losses relative to the unclipped twin.

No run significantly outperforms the base model (Figure~\ref{fig:aime25-row}a). K show only minimal accuracy changes, whereas F, L, and M degrade substantially, and most clipped runs perform worse than their unclipped twins. M is the exception: both of its run collapse, and its twin reads zero only because it stops closing the thinking tag, so the strict scorer finds no answer.

Terminal loop rate separates the runs more sharply (Figure~\ref{fig:aime25-row}b). Every clipped run ends in a terminal loop far more often than the base model, while every unclipped twin stays near the base rate. Looped responses fail to reach a final answer and are
therefore scored as incorrect. Accuracy losses in the clipped runs
follow the same ordering as their loop rates: smallest in K,
intermediate in F, and largest in L and M.

With 12 samples per problem, no clipped endpoint has lower pass@12
than its twin on either benchmark: every problem solved by the twin
is also solved at least once by the clipped arm. We interpret this
pattern as evidence that clipped training reduces per-sample
reliability. However, the unclipped F and L twins remain below the
fixed-base mean despite rarely looping. Thus, clipping-induced
looping contributes to the observed degradation, but does not
explain all accuracy loss relative to the base model.

Appendix~\ref{app:evaluation} gives avg@12, terminal loop rate, and pass@12 at every checkpoint on AIME 2024 and AIME 2025 benchmarks.

\section{Conclusion}
Pointwise clipping was introduced to stabilize training against dominant stylistic
tokens, but it also changes the direction in which forward kl corrects the student.
In training this change appears as repetitions that continue to the end of the
response, a loss of training stability. Because this failure develops over updates,
studies that train with the pointwise-clipped objective could report accuracy and a 
text degeneration measure, across checkpoints, in addition
to their best checkpoint. We hope that understanding these clipping dynamics helps
make on-policy self-distillation stable enough for its gains to persist over longer
training.

\bibliography{iclr2027_conference}
\bibliographystyle{iclr2027_conference}

\appendix
\section{Proof of Proposition~\ref{prop:piecewise-gradient}.}
\label{app:proofs}

Fix a response position and suppress its index $t$. Hold the finite support
$\mathcal S$ and the teacher distribution $p$ fixed. By
Equation~\ref{eq:support-distributions}, $p_j,q_j>0$ for every
$j\in\mathcal S$, and both distributions sum to one on $\mathcal S$.
For the temperature-scaled student logits $z_j=b_j/T$, write
\[
Z=\sum_{j\in\mathcal S}\exp(z_j),
\qquad
q_j=\frac{\exp(z_j)}{Z}.
\]

Consider any logit vector $z^\circ$ satisfying $d_j\neq\tau$ for all
$j\in\mathcal S$, and let $A$ and $C$ be its active and clipped sets as
defined in Section~\ref{sec:local}.
Each $d_j=p_j\log(p_j/q_j)$ is smooth in $z$, since $p_j$ is
fixed and $q_j$ is a positive smooth function of $z$.
Because $\mathcal S$ is finite and no $d_j$ equals $\tau$ at
$z^\circ$, there is an open neighborhood of $z^\circ$ on which
$A$ and $C$, and hence $P_A=\sum_{j\in A}p_j$, remain unchanged.
On this neighborhood, Equation~\ref{eq:clip-region} gives
\[
D_{\mathrm{clip}}=\sum_{j\in A}d_j+|C|\tau,
\]
with $A$ and $C$ fixed at their values at $z^\circ$.
This expression is smooth in $z$, so $D_{\mathrm{clip}}$ is
differentiable on the neighborhood and its gradient is obtained
by differentiating the displayed expression.

For any $i,j\in\mathcal S$, the softmax identity
$\log q_j=z_j-\log Z$ gives
\[
d_j=p_j\log p_j-p_jz_j+p_j\log Z.
\]
Because $p$ is fixed and
\[
\frac{\partial z_j}{\partial z_i}=\mathbf 1\{j=i\},
\qquad
\frac{\partial\log Z}{\partial z_i}
=\frac{1}{Z}\frac{\partial Z}{\partial z_i}
=\frac{\exp(z_i)}{Z}=q_i,
\]
it follows that
\begin{equation}
\frac{\partial d_j}{\partial z_i}
=-p_j\mathbf 1\{j=i\}+p_jq_i.
\label{eq:appendix-term-derivative}
\end{equation}

Since $D_{\mathrm{FKL}}=\sum_{j\in\mathcal S}d_j$ by
Equation~\ref{eq:losses}, summing
Equation~\ref{eq:appendix-term-derivative} over $\mathcal S$
gives
\begin{align*}
\frac{\partial D_{\mathrm{FKL}}}{\partial z_i}
&=\sum_{j\in\mathcal S}
  \bigl(-p_j\mathbf 1\{j=i\}+p_jq_i\bigr)\\
&=-p_i+q_i\sum_{j\in\mathcal S}p_j
=q_i-p_i,
\end{align*}
at every logit vector, including points on the clipping boundaries.
For the clipped objective, $|C|\tau$ has zero derivative on the
neighborhood under consideration. Summing only over $A$ therefore gives
\begin{align}
\frac{\partial D_{\mathrm{clip}}}{\partial z_i}
&=\sum_{j\in A}
  \bigl(-p_j\mathbf 1\{j=i\}+p_jq_i\bigr)\notag\\
&=-p_i\mathbf 1\{i\in A\}+q_i\sum_{j\in A}p_j\notag\\
&=P_Aq_i-p_i\mathbf 1\{i\in A\}.
\label{eq:appendix-gradient}
\end{align}
This establishes Equation~\ref{eq:piecewise-gradient}.

Now let $i\in C$. Then $d_i>\tau>0$, and $p_i>0$ implies
\[
\log\frac{p_i}{q_i}>0,
\qquad\text{hence}\qquad p_i>q_i.
\]
Consequently, $\partial D_{\mathrm{FKL}}/\partial z_i=q_i-p_i<0$.
Since $i\notin A$, Equation~\ref{eq:appendix-gradient} reduces to
$\partial D_{\mathrm{clip}}/\partial z_i=P_Aq_i$.
To establish strict positivity, first note that $q_i>0$. Moreover,
$\sum_{j\in\mathcal S}q_j=\sum_{j\in\mathcal S}p_j$ implies the existence
of a token $j$ with $q_j\ge p_j$. For this token,
\[
d_j=p_j\log\frac{p_j}{q_j}\le0<\tau,
\]
so $j\in A$ and $P_A\ge p_j>0$. Therefore,
\[
\frac{\partial D_{\mathrm{FKL}}}{\partial z_i}=q_i-p_i<0,
\qquad
\frac{\partial D_{\mathrm{clip}}}{\partial z_i}=P_Aq_i>0,
\]
which proves Equation~\ref{eq:logit-gradient-reversal}. Since $z^\circ$
was arbitrary away from the clipping boundaries, the result holds at
every point specified in the proposition. \qed

\paragraph{A simpler example of updating dynamics}

To understand the direction of the updates on the logits as suggested by the previous proposition, the example below can give some intuitions. In the most general case, decreasing certain logits on index $i$ does not automatically mean to decrease the corresponding probability $q_i$--but in the case when other $z$ values are fixed, this holds true. We can construct a continuous-time dynamical system to mimic the optimization process on the objective functions of the two model versions, defining the parameter updates as the negative gradient of the loss with respect to the logits ($\frac{dz_i}{dt} = - \nabla_{z_i} D$). For the exact forward kl objective $D_{FKL}$, this logit update is defined continuously as:

$$\frac{dz_i}{dt}(t) = p_i - q_i(t)$$

For the clipped objective, the gradient space is instead split into a piecewise system governed by the local divergence threshold $d_j$:

$$\frac{dz_i}{dt}(t) = \begin{cases} p_i - P_A q_i(t), & \text{if } d_j \leq \tau \\ -P_A q_i(t), & \text{if } d_j > \tau \end{cases}$$

\begin{figure}
    \centering
    \includegraphics[width=1\linewidth]{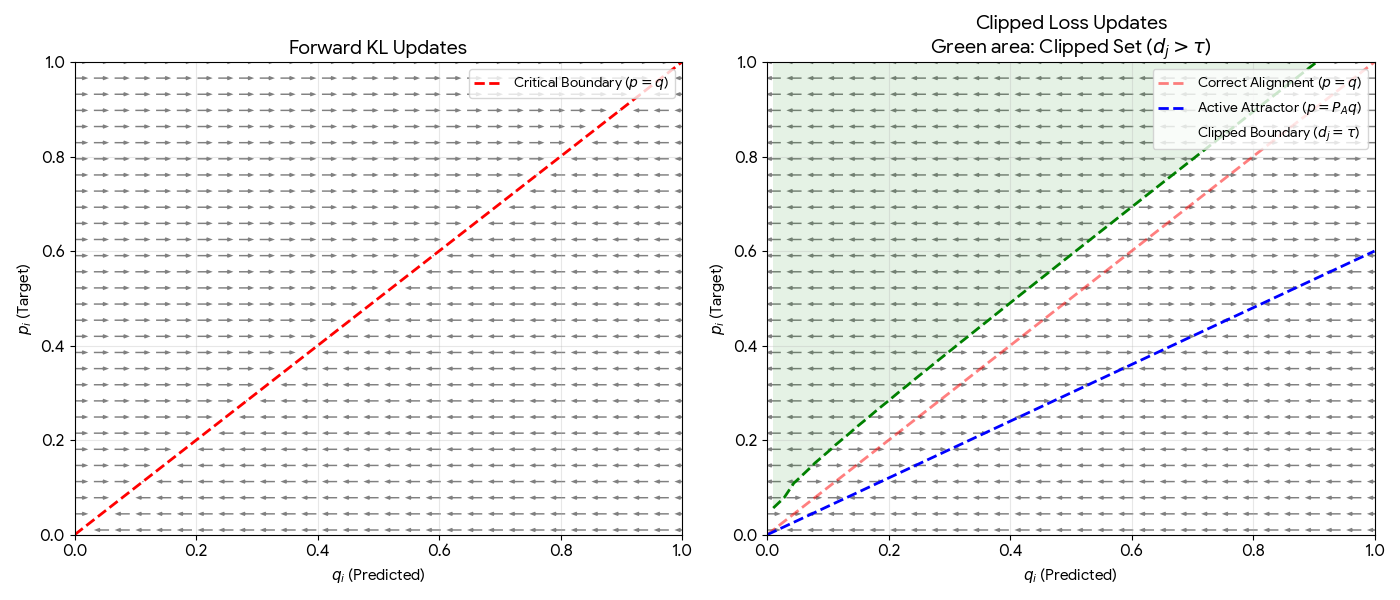}
    \caption{A simple example of updating direction of q under different losses(Forward KL v.s. Clipped Loss Updates.}
    \label{fig:loss_alignment_example}
\end{figure}

To map these raw logit updates $\frac{dz_i}{dt}$ to the resulting probability trajectories $\frac{dq_i}{dt}$, we must account for the coupling effect of the softmax function, where the evolution of a single probability $q_i$ depends on the updates to all logits in the vocabulary. However, to isolate the localized forces acting on a single prediction and visualize them within a simplified 2D direction field diagram, we introduce a specific framing assumption where all other logits $z_j, j\neq i$ are temporarily held constant ($\frac{dz_j}{dt} = 0$). Under this isolated assumption, because the softmax function is strictly monotonically increasing with respect to its own logit, $\frac{dq_i}{dt}$ acts as a positive monotonic scaling of $\frac{dz_i}{dt}$. This strict sign preservation guarantees that the direction of the probability update exactly mirrors the direction of the logit update, providing the mathematical justification to map the negative gradients directly onto a 2D phase portrait to evaluate their critical points and structural stability. The direction field, null-cline, contour for the clipping region and the diagonal (for the correct alignment p=q) is shown in figure \ref{fig:loss_alignment_example}.

Applying this localized framework to the exact forward kl divergence (left) demonstrates that the gradient acts as a proportional restorative force across the entire probability space, driving the system toward a single, globally stable equilibrium along the critical boundary where $p_i = q_i$ (the diagonal). Because the directional derivative relies entirely on the unaltered residual, the vector field smoothly and pushes the predictions directly toward perfect calibration. This ensures the parameter space remains entirely free of dead zones, structural barriers, or competing attractors.

Conversely, the clipped objective (right) shatters the global stability of the forward kl. In the active set (unshaded area on bottom-right corner) where the divergence is low ($d_j < \tau$), scaling the predicted term by $P_A$ artificially shifts the fixed point away from true alignment, forcing the vectors toward a false attractor along the line $p_i = P_A q_i$ (the blue line with slope that is less steep). When the local divergence exceeds the $\tau$ threshold, the system crosses into the clipped set (green shaded area on the top left), completely erasing the target probability $q_i$ from the derivative calculation. Inside this boundary, the dynamic devolves into pure probability decay. Because the gradient equation yields a negative update ($-P_A q_i$), it relentlessly pushes the predictions until $0$ is reached, structurally preventing the system from ever recovering true calibration when initialized or driven too far from the target. To sum up, the green shaded area along with the area between the red and blue lines on the right are where the arrows for the clipped objectives are in a different direction as compared with the ones for forward KL. One might observe that if you trace a phase portrait alone the arrows, then if you start with a point in either clipped or active set, the trajectory never goes into the other--this may give some intuition for boundary check for the next proposition as well.

\section{Proof of \texorpdfstring{Proposition~\ref{prop:support-pruned}}{Proposition 2}}
\label{app:proof-prop2}

\paragraph{Setup.}
Fix one response position. Let $p$ be the teacher distribution on the finite candidate
set $\mathcal S$. It is a softmax output, so $p_i>0$ for every $i\in \mathcal S$, and $\sum_{i\in \mathcal S}p_i=1$.
Let $\tau>0$.
Student distributions are the points of the closed simplex,
\[
  q_i\ge0\quad(i\in \mathcal S),\qquad\sum_{i\in \mathcal S}q_i=1,
\]
so probabilities equal to zero are allowed. We use the convention
$p_j\log(p_j/0)=+\infty$. The capped term of a zero-probability token is then
$\min\{+\infty,\tau\}=\tau$, so
\[
  D_{\mathrm{clip}}(q)=\sum_{j\in \mathcal S}\min\Bigl\{p_j\log\frac{p_j}{q_j},\ \tau\Bigr\}
\]
is finite on the whole simplex.

Let $A$ and $C$ be fixed nonempty disjoint sets with $A\cup C=\mathcal S$. The \emph{fixed
clipping region} $R$ is the set of student distributions $q$ with
\[
  p_i\log\frac{p_i}{q_i}\le\tau\quad(i\in A),
  \qquad
  p_j\log\frac{p_j}{q_j}>\tau\quad(j\in C).
\]
Write $P_A=\sum_{i\in A}p_i$ and $Q_C=\sum_{j\in C}q_j$. Since $A$ and $C$ are nonempty
and every $p_i>0$, we have $0<P_A<1$. For $q\in R$, every active term is at most $\tau$,
so its minimum with $\tau$ is the term itself, and every clipped term exceeds $\tau$, so
its minimum with $\tau$ is $\tau$. Hence, for $q\in R$ (Equation~\ref{eq:clip-region}),
\[
  D_{\mathrm{clip}}(q)=\sum_{i\in A}p_i\log\frac{p_i}{q_i}+\sum_{j\in C}\tau
  =F(q_A)+|C|\tau,
  \qquad F(q_A)=\sum_{i\in A}p_i\log\frac{p_i}{q_i},
  \qquad q_A=(q_i)_{i\in A}.
\]
This identity is valid on $R$. Outside $R$ it fails in general, because a term that
exceeds $\tau$ is capped.

\noindent\textbf{Claim.} The distribution $q^\star$ with $q^\star_i=p_i/P_A$ for $i\in A$ and
$q^\star_j=0$ for $j\in C$ is the unique minimizer of $D_{\mathrm{clip}}$ over $R$, and
$D_{\mathrm{clip}}(q^\star)=P_A\log P_A+|C|\tau$ (Equation~\ref{eq:fixed-region-minimizer}).

\paragraph{Overview.}
The proof has two steps. \emph{Step~1} settles how much mass sits on the clipped set:
any such mass can be moved onto an active token without leaving $R$, and doing so
strictly lowers the loss, so only distributions with $Q_C=0$ can compete. \emph{Step~2}
settles how that mass is split within the active set. Dropping the active-token clipping
inequalities leaves a larger, smooth problem: minimize the strictly convex function $F$
over the positive allocations with $\sum_{i\in A}q_i=1$. Its unique global minimizer is
the Lagrange point $q_i=p_i/P_A$, which exceeds $p_i$ because $P_A<1$. \emph{The crucial
implication is that the unique global minimizer of the larger, smooth problem satisfies
the original clipping constraints.} It is therefore also the unique minimizer of the
original, restricted problem. The larger problem is only an auxiliary device: outside $R$
the formula $F$ is not the clipped objective, and nothing is claimed there.

\paragraph{Membership facts (M1)--(M4).}
Steps~1 and~2 use four facts about when a token is active or clipped. These are the only
places where the threshold enters the proof. Fix a token $i$ and regard its term
$p_i\log(p_i/x)$ as a function of its student probability $x\ge0$, with value $+\infty$
at $x=0$. For $x>0$,
\[
  \frac{d}{dx}\Bigl[p_i\log\frac{p_i}{x}\Bigr]
  =\frac{d}{dx}\bigl[p_i\log p_i-p_i\log x\bigr]
  =-\frac{p_i}{x}<0 ,
\]
so the term is strictly decreasing in $x$ on $(0,\infty)$, and at $x=p_i$ it equals
$p_i\log1=0$.
\begin{enumerate}[label=(M\arabic*),leftmargin=3.2em,itemsep=2pt,topsep=3pt]
  \item \emph{Every active token has $q_i>0$.} If $q_i=0$, the term is $+\infty>\tau$, so
        token $i$ would be clipped, not active. Consequently $F$ is finite on $R$.
  \item \emph{An active token that gains probability stays active.} If $q'_i\ge q_i>0$,
        then, because the term is decreasing,
        $p_i\log(p_i/q'_i)\le p_i\log(p_i/q_i)\le\tau$.
  \item \emph{A clipped token that loses probability stays clipped, including at zero.}
        Let $0\le q'_j\le q_j$. If $q'_j=0$, the term is $+\infty>\tau$. If $q'_j>0$, then
        $q_j>0$ as well and $p_j\log(p_j/q'_j)\ge p_j\log(p_j/q_j)>\tau$.
  \item \emph{A token with $q_i\ge p_i$ is active for every $\tau>0$.} Because the term is
        decreasing, $p_i\log(p_i/q_i)\le p_i\log(p_i/p_i)=0<\tau$.
\end{enumerate}

\paragraph{Step 1: mass on the clipped set prevents optimality.}
Let $q\in R$ with $Q_C>0$. Pick any active token $k\in A$ and move all clipped mass onto
it:
\[
  q'_k=q_k+Q_C,\qquad q'_j=0\ \ (j\in C),\qquad q'_i=q_i\ \ (i\in A,\ i\neq k).
\]
\emph{Normalization.}
$\sum_{i\in \mathcal S}q'_i=\sum_{i\in A,\,i\neq k}q_i+(q_k+Q_C)+0=\sum_{i\in A}q_i+Q_C=1$.

\noindent\emph{Same region.} Token $k$ gains probability, so it stays active by (M2).
Every clipped token drops to zero, so it stays clipped by (M3). The other active tokens
are unchanged. Hence $q'\in R$, and the identity $D_{\mathrm{clip}}=F+|C|\tau$ applies to both $q$
and $q'$.

\noindent\emph{Difference.} The constants $|C|\tau$ cancel, and so do all active terms
with $i\neq k$. What remains is the $k$-th term:
\begin{align*}
  D_{\mathrm{clip}}(q')-D_{\mathrm{clip}}(q)
  &=p_k\log\frac{p_k}{q_k+Q_C}-p_k\log\frac{p_k}{q_k}\\
  &=p_k\bigl[\log q_k-\log(q_k+Q_C)\bigr]
  =p_k\log\frac{q_k}{q_k+Q_C}.
\end{align*}
Here $p_k$ and $q_k$ are the teacher and student probabilities of the single receiving
token $k$, not the set masses. We have $p_k>0$, $q_k>0$ by (M1), and $Q_C>0$, so
$0<q_k/(q_k+Q_C)<1$, the logarithm is negative, and
\[
  D_{\mathrm{clip}}(q')<D_{\mathrm{clip}}(q).
\]
Hence every distribution in $R$ with $Q_C>0$ has strictly larger loss than some
distribution in $R$ with $Q_C=0$.

\paragraph{Step 2: optimal allocation on the active set.}
Throughout this step $p_i>0$ for every $i$, because the teacher is a softmax output, and
$q_i>0$ for every $i\in A$ by (M1). Neither is an extra assumption. Consequently $F$ is
finite and smooth wherever it is evaluated, and every Hessian entry $p_i/q_i^{2}$ below is
strictly positive.

\emph{(a) The restricted problem.} After Step~1 the competitors are the distributions
$q\in R$ with $Q_C=0$. For such $q$ we have $q_j=0$ on $C$, $q_i>0$ on $A$,
$\sum_{i\in A}q_i=1$, and $D_{\mathrm{clip}}(q)=F(q_A)+|C|\tau$. So we must minimize $F$ over
\[
  K_\tau=\Bigl\{q_A\in\mathbb{R}^A:\ q_i>0,\ \ \sum_{i\in A}q_i=1,\ \
  p_i\log\frac{p_i}{q_i}\le\tau\ \ \forall i\in A\Bigr\}.
\]
Conversely, every $q_A\in K_\tau$, extended by zeros on $C$, is a distribution in $R$
with $Q_C=0$.

\emph{(b) The relaxed problem.} Drop the threshold inequalities and let
\[
  K=\Bigl\{q_A\in\mathbb{R}^A:\ q_i>0,\ \ \sum_{i\in A}q_i=1\Bigr\}\ \supseteq\ K_\tau .
\]
We minimize the same formula $F$ over $K$. On $K\setminus K_\tau$ the formula $F$ is no
longer the clipped objective, because there some term exceeds $\tau$ and would be capped.
The relaxed problem is an auxiliary smooth problem. It enters the argument only through
the inclusion $K_\tau\subseteq K$, and nothing is claimed about $D_{\mathrm{clip}}$ outside $R$.

\emph{(c) Derivatives of $F$.} We differentiate on the positive orthant
$\{q_A:\ q_i>0\}$, which is open, convex, and contains $K$. Write
\[
  F(q_A)=\sum_{i\in A}\bigl[p_i\log p_i-p_i\log q_i\bigr].
\]
Only the $i$-th summand depends on $q_i$, and $p_i\log p_i$ is a constant, so
\[
  \frac{\partial F}{\partial q_i}
  =\frac{\partial}{\partial q_i}\bigl[p_i\log p_i-p_i\log q_i\bigr]
  =0-p_i\cdot\frac{1}{q_i}
  =-\frac{p_i}{q_i}.
\]
For the second derivatives, differentiate $-p_i/q_i=-p_i\,q_i^{-1}$ with respect to
$q_m$. If $m=i$,
\[
  \frac{\partial^{2}F}{\partial q_i^{2}}
  =\frac{\partial}{\partial q_i}\bigl[-p_i\,q_i^{-1}\bigr]
  =-p_i\cdot(-1)\,q_i^{-2}
  =\frac{p_i}{q_i^{2}} .
\]
If $m\neq i$, the expression $-p_i/q_i$ does not depend on $q_m$, so
$\partial^{2}F/\partial q_i\,\partial q_m=0$. The Hessian is therefore diagonal,
$\nabla^{2}F(q_A)=\mathrm{diag}\bigl(p_i/q_i^{2}\bigr)_{i\in A}$, and for every vector
$v\in\mathbb{R}^A$,
\[
  v^{\top}\nabla^{2}F(q_A)\,v
  =\sum_{i\in A}\sum_{m\in A}v_i\,\frac{\partial^{2}F}{\partial q_i\,\partial q_m}\,v_m
  =\sum_{i\in A}\frac{p_i}{q_i^{2}}\,v_i^{2}.
\]
Every coefficient $p_i/q_i^{2}$ is strictly positive, because $p_i>0$ and $q_i>0$. If
$v\neq0$, some $v_i\neq0$, so the sum is strictly positive: the Hessian is positive
definite at every point of the orthant.

\emph{(d) Strict convexity.} A function $f$ on a convex set is \emph{strictly convex} if
$f(\theta x+(1-\theta)y)<\theta f(x)+(1-\theta)f(y)$ for all $x\neq y$ in the set and all
$\theta\in(0,1)$. We use two standard facts. (F1)~A twice differentiable function on an
open convex set whose Hessian is positive definite at every point is strictly convex.
(F2)~A differentiable strictly convex function lies strictly above each of its tangent
planes: $f(y)>f(x)+\nabla f(x)^{\top}(y-x)$ for all $x\neq y$. By (c) and (F1), $F$ is
strictly convex on the orthant, independently of normalization, and hence on its convex
subset $K$.

\emph{(e) Lagrange point.} Introduce a multiplier $\lambda$ for the normalization
constraint:
\[
  \mathcal{L}(q_A,\lambda)=F(q_A)+\lambda\Bigl(\sum_{m\in A}q_m-1\Bigr).
\]
Using (c) and $\frac{\partial}{\partial q_i}\bigl(\sum_{m\in A}q_m-1\bigr)=1$,
\[
  \frac{\partial\mathcal{L}}{\partial q_i}=-\frac{p_i}{q_i}+\lambda=0
  \quad\Longrightarrow\quad
  \lambda=\frac{p_i}{q_i}
  \quad\Longrightarrow\quad
  q_i=\frac{p_i}{\lambda}\qquad\text{for every }i\in A .
\]
So the ratio $p_i/q_i$ is the same for every active token. Substituting into the
normalization constraint $\partial\mathcal{L}/\partial\lambda=\sum_{i\in A}q_i-1=0$,
\[
  1=\sum_{i\in A}\frac{p_i}{\lambda}=\frac{1}{\lambda}\sum_{i\in A}p_i=\frac{P_A}{\lambda}
  \quad\Longrightarrow\quad
  \lambda=P_A
  \quad\Longrightarrow\quad
  q^\star_i=\frac{p_i}{P_A}\qquad(i\in A).
\]
This is the only stationary point, since the equations force $q_i=p_i/\lambda$ and then
$\lambda=P_A$. It lies in $K$: $q^\star_i>0$ and $\sum_{i\in A}q^\star_i=P_A/P_A=1$.

\emph{(f) Global minimality in the relaxed problem.} The gradient at the Lagrange point
is
\[
  \frac{\partial F}{\partial q_i}\Big|_{q^\star_A}
  =-\frac{p_i}{q^\star_i}=-\frac{p_i}{p_i/P_A}=-P_A
  \quad\text{for every }i\in A,
  \qquad\text{that is,}\qquad \nabla F(q^\star_A)=-P_A\mathbf{1}.
\]
Let $q_A\in K$ with $q_A\neq q^\star_A$. By (F2) with $x=q^\star_A$ and $y=q_A$,
\[
  F(q_A)>F(q^\star_A)+\nabla F(q^\star_A)^{\top}(q_A-q^\star_A),
\]
and the last term vanishes because both allocations sum to one:
\[
  \nabla F(q^\star_A)^{\top}(q_A-q^\star_A)
  =\sum_{i\in A}(-P_A)(q_i-q^\star_i)
  =-P_A\Bigl(\sum_{i\in A}q_i-\sum_{i\in A}q^\star_i\Bigr)
  =-P_A(1-1)=0 .
\]
Hence $F(q_A)>F(q^\star_A)$ for every $q_A\in K$ other than $q^\star_A$: the Lagrange point is
the unique global minimizer of the relaxed problem.

\emph{(g) Membership of the candidate.} Since $0<P_A<1$, we have
$q^\star_i=p_i/P_A>p_i$ for every $i\in A$, and
\[
  p_i\log\frac{p_i}{q^\star_i}=p_i\log\frac{p_i}{p_i/P_A}=p_i\log P_A<0<\tau ,
\]
in agreement with (M4). Every active term lies strictly below the threshold, for every
$\tau>0$, so the candidate cannot cross into the clipped set. Thus $q^\star_A\in K_\tau$, and with
$q^\star_j=0$ on $C$ we get $q^\star\in R$.

\emph{(h) Back to the restricted problem.} Let $q_A\in K_\tau$ with $q_A\neq q^\star_A$. Since
$K_\tau\subseteq K$, part~(f) gives $F(q_A)>F(q^\star_A)$, and $q^\star_A\in K_\tau$ by~(g). So
$q^\star_A$ is also the unique global minimizer of the restricted problem: a minimizer over
the larger set that lies in the smaller set is the minimizer over the smaller set. Its
value is
\[
  F(q^\star_A)=\sum_{i\in A}p_i\log\frac{p_i}{p_i/P_A}
  =\sum_{i\in A}p_i\log P_A
  =\Bigl(\sum_{i\in A}p_i\Bigr)\log P_A
  =P_A\log P_A .
\]

\paragraph{Conclusion.}
Let $q\in R$ with $q\neq q^\star$.
\emph{Case $Q_C>0$.} Step~1 gives $q'\in R$ with $Q_C=0$ and $D_{\mathrm{clip}}(q)>D_{\mathrm{clip}}(q')$, and
Step~2 gives $D_{\mathrm{clip}}(q')=F(q'_A)+|C|\tau\ge F(q^\star_A)+|C|\tau=D_{\mathrm{clip}}(q^\star)$. So
$D_{\mathrm{clip}}(q)>D_{\mathrm{clip}}(q^\star)$.
\emph{Case $Q_C=0$.} Then $q_j=0=q^\star_j$ on $C$, so $q\neq q^\star$ means $q_A\neq q^\star_A$, and
Step~2 gives $D_{\mathrm{clip}}(q)=F(q_A)+|C|\tau>F(q^\star_A)+|C|\tau=D_{\mathrm{clip}}(q^\star)$.

\noindent In both cases $D_{\mathrm{clip}}(q)>D_{\mathrm{clip}}(q^\star)$. Hence $q^\star$ is the unique minimizer of
$D_{\mathrm{clip}}$ over $R$, with $D_{\mathrm{clip}}(q^\star)=P_A\log P_A+|C|\tau$. Existence is not assumed: the
minimizer is exhibited. \qed

\paragraph{Fixed $Q_C$ (Equation~\ref{eq:support-pruned}).}
Let $c$ be a value of $Q_C$ attained in $R$. For $q\in R$ with $Q_C=c$, the loss is
$F(q_A)+|C|\tau$ with $q_i>0$ and $\sum_{i\in A}q_i=1-c$, and it does not depend on how
$c$ is split among the clipped tokens as long as it does not make any one of them active by going over the $d>\tau$ restriction. Repeat Step~2 with the constraint
$\sum_{i\in A}q_i=1-c$. Stationarity again gives $q_i=p_i/\lambda$, and now
\[
  1-c=\sum_{i\in A}\frac{p_i}{\lambda}=\frac{P_A}{\lambda}
  \quad\Longrightarrow\quad
  \lambda=\frac{P_A}{1-c}
  \quad\Longrightarrow\quad
  q^\star_i(c)=\frac{1-c}{P_A}\,p_i .
\]
The gradient there is $-p_i/q^\star_i(c)=-P_A/(1-c)$ in every coordinate, so for any other
positive allocation with the same sum,
$\nabla F(q^\star_A(c))^{\top}(q_A-q^\star_A(c))=-\frac{P_A}{1-c}\bigl((1-c)-(1-c)\bigr)=0$, and
(F2) gives $F(q_A)>F(q^\star_A(c))$: this is the unique global minimizer of the relaxed
problem at level $c$.

\emph{Membership.} Every clipped token has a term above $\tau>0$, so
$\log(p_j/q_j)>0$ and $q_j<p_j$. Summing over $C$ gives
$c<\sum_{j\in C}p_j=1-P_A$, hence $(1-c)/P_A>1$ and $q^\star_i(c)>p_i$. By (M4) every active
term is below $\tau$, so the relaxed minimizer satisfies the threshold inequalities and is
also the minimizer of the restricted problem at level $c$. Its loss is
\begin{align*}
  D^{\star}_{\mathrm{clip}}(c)
  &=\sum_{i\in A}p_i\log\frac{p_i}{(1-c)\,p_i/P_A}+|C|\tau\\
  &=P_A\log\frac{P_A}{1-c}+|C|\tau
  =P_A\log P_A-P_A\log(1-c)+|C|\tau ,
\end{align*}
and
\[
  \frac{dD^{\star}_{\mathrm{clip}}}{dc}
  =-P_A\cdot\frac{-1}{1-c}=\frac{P_A}{1-c}>0 .
\]
The level-$c$ minimizer is unique in its active coordinates only.

\paragraph{Remarks.}
\emph{(i) $F$ is not a KL divergence.} Its arguments $p_A$ and $q_A$ are sub-probability
vectors, and $F$ can be negative: $F(q^\star_A)=P_A\log P_A<0$. The proof uses only the strict
convexity of $F$.

\emph{(ii) Scope.} The statement concerns one fixed clipping region; no comparison across
clipping patterns is claimed. A softmax over finite logits assigns every token positive
probability, so such a student never equals $q^\star$; by the fixed-$Q_C$ statement the
optimal loss at level $c$ decreases to $D_{\mathrm{clip}}(q^\star)$ as $c\downarrow0$, so $q^\star$ is
approached but not attained.

\section{Training and Evaluation Configuration, Capture}
\label{app:methods}

\paragraph{Training configuration}
\label{app:training-details}
In all eight runs, the student and the teacher are initialized from the same
Qwen3-4B checkpoint.
Each clipped/unclipped pair receives the same per-update problems. 
Table~\ref{tab:training-recipe} lists the detailed training configuration used in our experiment.

\begin{table*}[t]
\centering
\small
\caption{Training Configuration.}
\label{tab:training-recipe}
\begin{tabular}{@{}p{0.21\textwidth}p{0.73\textwidth}@{}}
\toprule
Component & Executed value \\
\midrule
Base checkpoint & \texttt{Qwen/Qwen3-4B} \\
Optimizer & \texttt{torch.optim.AdamW}; learning rate $10^{-6}$; $\beta=(0.9,0.999)$; $\epsilon=10^{-8}$; weight decay $0.01$ \\
Schedule & Zero warmup. $10^{-6}$ constant. \\
\raggedright Updates and checkpoints & 200 updates; a checkpoint is saved every 25 updates \\
Seed & Training and data-loader seeds are 42 \\
Global update & 30 problems, one rollout each \\
Prompt length cap & 2,048 tokens; the response caps are given in Table~\ref{tab:run-matrix} \\
\raggedright Training rollout sampler & Temperature $1.1$, top-$p=0.8$, top-$k=20$ \\
Teacher sampler & The teacher returns its top-128 token log probabilities at temperature $1.0$ \\
Loss & Forward KL on the teacher's top-128 tokens, both distributions renormalized on that support at temperature $T=1.1$; clipped runs use $\tau=0.05$, unclipped runs apply no clip \\
Hardware & Four H200 GPUs per run: three for the student's training and rollouts, one for the dedicated SGLang teacher \\
Software & Transformers 4.57.1; PyTorch 2.9.1+cu128; Ray 2.54.1; SGLang 0.5.9; CUDA toolkit 12.8.2 \\
\bottomrule
\end{tabular}
\end{table*}

\paragraph{Message templates.}
We give the user messages and the chat template used to serialize the student and teacher prompts. The student's user message is identical across all eight runs. Its serialized prompt differs only by the empty thinking block described below. Before chat serialization, the student's single user message is:
\begin{quote}
\small\raggedright\ttfamily
Problem: \{problem\}\\
\mbox{}\\
Please reason step by step, and put your final answer within \textbackslash boxed\{\}.
\end{quote}
The teacher's single user message is:
\begin{quote}
\small\raggedright\ttfamily
Problem: \{problem\}  Here is a reference solution to this problem: === Reference Solution Begin === \{answer\} === Reference Solution End ===  After reading the reference solution above, make sure you truly understand the reasoning behind each step - do not copy or paraphrase it. Now, using your own words and independent reasoning, derive the same final answer to the problem above. Please reason step by step, and put your final answer within \textbackslash boxed\{\}.
\end{quote}
The placeholder \{answer\} is filled with the full reference solution from the dataset, not the final answer alone. Qwen serializes the one-message prompt as
\begin{quote}
\small\raggedright\ttfamily
<|im\_start|>user\\
\{content\}<|im\_end|>\\
<|im\_start|>assistant
\end{quote}
with no system or tool message. When thinking is disabled, the serialization additionally appends an empty \texttt{<think>} block followed by \texttt{</think>} and a blank line. The student's sampled response token IDs are appended to the teacher prompt before teacher scoring.

\paragraph{Capture setting.}
\label{app:capture}
At every response position of all 200 updates of all eight runs, the trainer stores the token the student emitted, the IDs of the teacher's top-128 candidate tokens, the teacher's log probabilities on those tokens, and the student's log probabilities on the same tokens, gathered from its full-vocabulary log-softmax in the forward pass that computes the loss.

\section{Complete AIME2024 and AIME2025 Evaluation}
\label{app:evaluation}

\paragraph{Evaluation configuration.}
\label{app:eval-protocol}
Checkpoints 25, 50, 75, 100, 125, 150, 175, and 200 are evaluated on AIME~2024 and AIME~2025. Each benchmark contains 30 problems.

Each saved checkpoint is evaluated with thinking enabled, drawing 12 samples per problem at temperature $.6$, top-$p=.95$, top-$k=20$, and minimum-$p=0$, with a 38,912-token output budget. Scoring uses only the text after the last closing thinking tag: Math-Verify \citep{Kydlicek_Math-Verify_Math_Verification} compares it with the reference answer, and if that comparison fails, the last \texttt{\textbackslash boxed\{\}} expression is compared with the reference answer as a string or a number. We evaluate the same base-model checkpoint eight times for both benchmarks to measure variation from response sampling.

\begin{figure*}[h]
\centering
\includegraphics[width=\textwidth]{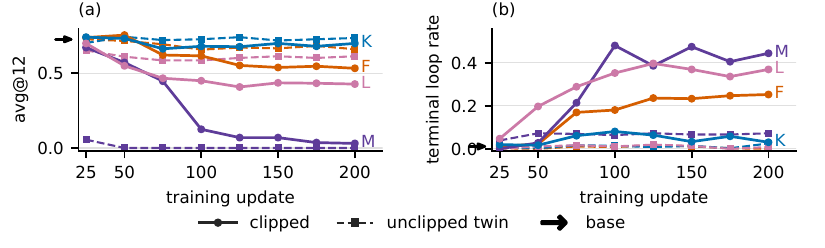}
\caption{AIME~2024 across training, in the layout of Figure~\ref{fig:aime25-row}. (a) Avg@12, the mean correct ratio over 12 samples for each of 30 problems at the checkpoint saved after each 25 updates. (b) The ratio of the 360 responses per checkpoint that end in a terminal periodic loop. Solid curves with circles are the clipped runs, dashed curves with squares their unclipped twins. Each trained curve is one realized trajectory. The black arrow on each vertical axis marks the base model, averaged over eight evaluation-sampling seeds. The values appear in Table~\ref{tab:aime24}.}
\label{fig:aime24-app}
\end{figure*}

\begin{table*}[!htb]
\centering
\scriptsize
\caption{Complete AIME~2025 evaluation at every checkpoint (columns: training update). (a) Avg@12, the mean correct ratio over 12 samples for each of 30 problems. (b) Terminal loop rate, the ratio of the 360 responses that end in a terminal periodic loop. (c) Pass@12, the ratio of the 30 problems solved by at least one of the 12 samples. Each trained row is one realized trajectory. The base row gives the mean and, in parentheses, the standard deviation over eight evaluation-sampling seeds of the fixed starting checkpoint.}
\label{tab:aime25}
\begin{tabular*}{\textwidth}{@{\extracolsep{\fill}}lrrrrrrrr@{}}
\toprule
\multicolumn{9}{@{}l}{(a) avg@12} \\
\midrule
Run & 25 & 50 & 75 & 100 & 125 & 150 & 175 & 200 \\
\midrule
F & $.661$ & $.653$ & $.589$ & $.550$ & $.581$ & $.525$ & $.531$ & $.544$ \\
F-no-clip & $.622$ & $.644$ & $.578$ & $.544$ & $.619$ & $.561$ & $.575$ & $.572$ \\
M & $.592$ & $.481$ & $.378$ & $.122$ & $.050$ & $.067$ & $.031$ & $.022$ \\
M-no-clip & $.047$ & $.000$ & $.000$ & $.000$ & $.000$ & $.000$ & $.000$ & $.000$ \\
K & $.664$ & $.653$ & $.647$ & $.619$ & $.622$ & $.550$ & $.628$ & $.600$ \\
K-no-clip & $.647$ & $.642$ & $.633$ & $.611$ & $.644$ & $.642$ & $.608$ & $.608$ \\
L & $.636$ & $.525$ & $.533$ & $.486$ & $.469$ & $.439$ & $.469$ & $.458$ \\
L-no-clip & $.581$ & $.531$ & $.481$ & $.508$ & $.489$ & $.506$ & $.508$ & $.517$ \\
\midrule
Base & \multicolumn{8}{c}{$.645\;(.010)$} \\
\midrule
\multicolumn{9}{@{}l}{(b) terminal loop rate} \\
\midrule
Run & 25 & 50 & 75 & 100 & 125 & 150 & 175 & 200 \\
\midrule
F & $.008$ & $.033$ & $.128$ & $.161$ & $.153$ & $.203$ & $.194$ & $.192$ \\
F-no-clip & $.011$ & $.003$ & $.008$ & $.019$ & $.000$ & $.006$ & $.006$ & $.008$ \\
M & $.006$ & $.036$ & $.192$ & $.494$ & $.389$ & $.422$ & $.400$ & $.400$ \\
M-no-clip & $.050$ & $.039$ & $.036$ & $.044$ & $.031$ & $.053$ & $.042$ & $.017$ \\
K & $.014$ & $.033$ & $.036$ & $.050$ & $.058$ & $.078$ & $.086$ & $.061$ \\
K-no-clip & $.011$ & $.011$ & $.006$ & $.006$ & $.000$ & $.003$ & $.008$ & $.011$ \\
L & $.047$ & $.144$ & $.211$ & $.231$ & $.256$ & $.328$ & $.267$ & $.303$ \\
L-no-clip & $.003$ & $.014$ & $.019$ & $.006$ & $.011$ & $.006$ & $.022$ & $.011$ \\
\midrule
Base & \multicolumn{8}{c}{$.006\;(.003)$} \\
\midrule
\multicolumn{9}{@{}l}{(c) pass@12} \\
\midrule
Run & 25 & 50 & 75 & 100 & 125 & 150 & 175 & 200 \\
\midrule
F & $.800$ & $.800$ & $.867$ & $.800$ & $.833$ & $.833$ & $.867$ & $.800$ \\
F-no-clip & $.867$ & $.800$ & $.800$ & $.733$ & $.833$ & $.833$ & $.800$ & $.800$ \\
M & $.867$ & $.700$ & $.667$ & $.367$ & $.233$ & $.267$ & $.167$ & $.133$ \\
M-no-clip & $.133$ & $.000$ & $.000$ & $.000$ & $.000$ & $.000$ & $.000$ & $.000$ \\
K & $.833$ & $.800$ & $.833$ & $.833$ & $.867$ & $.800$ & $.833$ & $.833$ \\
K-no-clip & $.833$ & $.800$ & $.833$ & $.767$ & $.867$ & $.833$ & $.767$ & $.800$ \\
L & $.833$ & $.800$ & $.767$ & $.767$ & $.833$ & $.767$ & $.767$ & $.767$ \\
L-no-clip & $.767$ & $.767$ & $.733$ & $.767$ & $.767$ & $.767$ & $.700$ & $.767$ \\
\midrule
Base & \multicolumn{8}{c}{$.817\;(.018)$} \\
\bottomrule
\end{tabular*}
\label{tab:aime25-table}
\end{table*}

\begin{table*}[!htb]
\centering
\scriptsize
\caption{Complete AIME~2024 evaluation at every checkpoint (columns: training update). (a) Avg@12, the mean correct ratio over 12 samples for each of 30 problems. (b) Terminal loop rate, the ratio of the 360 responses that end in a terminal periodic loop. (c) Pass@12, the ratio of the 30 problems solved by at least one of the 12 samples. Each trained row is one realized trajectory. The base row gives the mean and, in parentheses, the standard deviation over eight evaluation-sampling seeds of the fixed starting checkpoint.}
\label{tab:aime24}
\begin{tabular*}{\textwidth}{@{\extracolsep{\fill}}lrrrrrrrr@{}}
\toprule
\multicolumn{9}{@{}l}{(a) avg@12} \\
\midrule
Run & 25 & 50 & 75 & 100 & 125 & 150 & 175 & 200 \\
\midrule
F & $.739$ & $.756$ & $.622$ & $.617$ & $.553$ & $.539$ & $.547$ & $.533$ \\
F-no-clip & $.728$ & $.717$ & $.694$ & $.658$ & $.672$ & $.667$ & $.683$ & $.661$ \\
M & $.672$ & $.572$ & $.447$ & $.125$ & $.069$ & $.069$ & $.036$ & $.031$ \\
M-no-clip & $.056$ & $.000$ & $.000$ & $.000$ & $.000$ & $.000$ & $.000$ & $.000$ \\
K & $.742$ & $.733$ & $.664$ & $.681$ & $.678$ & $.700$ & $.681$ & $.700$ \\
K-no-clip & $.703$ & $.744$ & $.722$ & $.728$ & $.742$ & $.719$ & $.728$ & $.736$ \\
L & $.700$ & $.550$ & $.467$ & $.450$ & $.408$ & $.436$ & $.433$ & $.428$ \\
L-no-clip & $.653$ & $.611$ & $.586$ & $.586$ & $.603$ & $.614$ & $.603$ & $.614$ \\
\midrule
Base & \multicolumn{8}{c}{$.727\;(.013)$} \\
\midrule
\multicolumn{9}{@{}l}{(b) terminal loop rate} \\
\midrule
Run & 25 & 50 & 75 & 100 & 125 & 150 & 175 & 200 \\
\midrule
F & $.011$ & $.022$ & $.169$ & $.181$ & $.236$ & $.233$ & $.247$ & $.253$ \\
F-no-clip & $.003$ & $.000$ & $.011$ & $.008$ & $.017$ & $.011$ & $.003$ & $.000$ \\
M & $.000$ & $.028$ & $.214$ & $.481$ & $.386$ & $.475$ & $.406$ & $.444$ \\
M-no-clip & $.039$ & $.072$ & $.069$ & $.064$ & $.072$ & $.067$ & $.067$ & $.072$ \\
K & $.019$ & $.017$ & $.061$ & $.081$ & $.064$ & $.033$ & $.058$ & $.031$ \\
K-no-clip & $.006$ & $.003$ & $.017$ & $.014$ & $.008$ & $.014$ & $.003$ & $.025$ \\
L & $.047$ & $.197$ & $.289$ & $.353$ & $.397$ & $.369$ & $.336$ & $.369$ \\
L-no-clip & $.000$ & $.011$ & $.017$ & $.011$ & $.019$ & $.014$ & $.000$ & $.008$ \\
\midrule
Base & \multicolumn{8}{c}{$.011\;(.003)$} \\
\midrule
\multicolumn{9}{@{}l}{(c) pass@12} \\
\midrule
Run & 25 & 50 & 75 & 100 & 125 & 150 & 175 & 200 \\
\midrule
F & $.867$ & $.833$ & $.900$ & $.867$ & $.900$ & $.867$ & $.867$ & $.833$ \\
F-no-clip & $.833$ & $.833$ & $.833$ & $.867$ & $.800$ & $.833$ & $.800$ & $.833$ \\
M & $.833$ & $.800$ & $.700$ & $.467$ & $.333$ & $.333$ & $.167$ & $.100$ \\
M-no-clip & $.233$ & $.000$ & $.000$ & $.000$ & $.000$ & $.000$ & $.000$ & $.000$ \\
K & $.833$ & $.867$ & $.833$ & $.833$ & $.867$ & $.867$ & $.833$ & $.867$ \\
K-no-clip & $.833$ & $.833$ & $.867$ & $.833$ & $.867$ & $.833$ & $.833$ & $.833$ \\
L & $.867$ & $.833$ & $.800$ & $.767$ & $.767$ & $.800$ & $.833$ & $.833$ \\
L-no-clip & $.833$ & $.833$ & $.800$ & $.800$ & $.800$ & $.800$ & $.800$ & $.800$ \\
\midrule
Base & \multicolumn{8}{c}{$.838\;(.033)$} \\
\bottomrule
\end{tabular*}
\label{tab:aime24-table}
\end{table*}

Figure ~\ref{fig:aime24-app} visualize AIME 2024 evaluation result: (a) avg@12 and, (b) terminal loop rate.  

Table ~\ref{tab:aime24-table} and  ~\ref{tab:aime25-table} shows the numerical value of per checkpoint evaluated on AIME 2024 and AIME 2025, respectively: (a) avg@12, (b) terminal loop rate and (c) pass@12.

\end{document}